\documentclass{article}

\input{packages.tex}
\newcommand{\dumarg}[1]	{\widetilde{#1}}
\newcommand{\bu}{\boldsymbol{u}}    						

\newcommand{\bv}{\boldsymbol{v}}    						
\newcommand{\bvpred}{\boldsymbol{v}}    				
\newcommand{\bvpreddot}{\dot{\boldsymbol{v}}}     

\newcommand{\dimrecon}{d_v}
\newcommand{\nv}{\dimrecon}    								
\newcommand{\bx}{\mathbf{x}}

\newcommand{\obs}{s}                   		 					
\newcommand{\bobs}{\boldsymbol{s}}
\newcommand{\bObs}{\boldsymbol{S}}
\newcommand{\nobs}{n_{\obs}}
\newcommand{\measope}{h}
\newcommand{\btheta}{\boldsymbol{\theta}}   			
\newcommand{\dumtheta}{\dumarg{\btheta}}
\newcommand{\ntheta}{n_{\boldsymbol{\theta}}}

\newcommand{\recmodel}{F_{\btheta}}
\newcommand{\recmodeldum}{F_{\dumarg{\btheta}}}
\newcommand{\recergo}{F_\mathrm{ergo}}
\newcommand{\recshort}{F_\mathrm{short}}
\newcommand{\recFP}{F_\mathrm{FP}}
\newcommand{\obstostate}{g}
\newcommand{\timedom}{\Gamma}
\newcommand{\phasespace}{\Omega}
\newcommand{\blatent}{\boldsymbol{\zeta}}
\newcommand{\regul}{r}
\newcommand{\loss}{\mathscr{L}}
\newcommand{\fp}{\bu_{\mathrm{FP}}}
\newcommand{\fpp}{\bu_{\mathrm{FP}_+}}
\newcommand{\fpm}{\bu_{\mathrm{FP}_-}}
\newcommand{\fpzero}{\bu_{\mathrm{FP}_0}}

\newcommand{\trainset}{\mathcal{T}}
\newcommand{\tsetrand}{\trainset_\mathrm{ens}}
\newcommand{\tsetergo}{\trainset_\mathrm{ergo}}
\newcommand{\tsetsplit}{\trainset_\mathrm{split}}
\newcommand{\tsetFP}{\trainset_\mathrm{FP}}
\newcommand{\tsetshort}{\trainset_\mathrm{short}}

\newcommand{\Dambient}{d} 
\newcommand{\attrac}{\boldsymbol{A}}
\newcommand{\dAttrac}{d_{\attrac}}
\newcommand{\dbu}{\Dambient}

\newcommand{\R}{\boldsymbol{R}}
\newcommand{\be}{\begin{equation}}
\newcommand{\ee}{\end{equation}}
\newcommand{\bea}{\begin{eqnarray}}
\newcommand{\eea}{\end{eqnarray}}
\newcommand{\benum}{\begin{enumerate}}
\newcommand{\eenum}{\end{enumerate}}
\newcommand{\bi}{\begin{itemize}}
\newcommand{\ei}{\end{itemize}}

\newcommand{\normLM}[2]{\left\|{#1}\right\|_{#2}}
\newcommand{\normsqLM}[2]{\left\|{#1}\right\|^2_{#2}}

\newcommand*{\arginf}	{{\mathrm{ arg\,inf}}}

\title{Curriculum learning for data-driven modeling of dynamical systems}

\author{Michele Alessandro Bucci \\ TAU--Team, INRIA Saclay \\ LISN, Universit\'e Paris-Saclay \\ 91190, Gif-sur-Yvette \\ France \\ \texttt{bucci.malessandro@gmail.com} \\ \And
Onofrio Semeraro \\ LISN-CNRS \\ Universit\'e Paris-Saclay \\ 91440, Orsay \\ France \\ \texttt{onofrio.semeraro@universite-paris-saclay.fr} \\ \And
Alexandre Allauzen \\ LAMSADE \\ Universit\'e Paris Dauphine \\75016 Paris \\ France \\ \texttt{alexandre.allauzen@dauphine.psl.eu} \\ \And
Sergio Chibbaro \\ LISN-CNRS \\ Universit\'e Paris-Saclay \\ 91440, Orsay \\ France \\ \texttt{sergio.chibbaro@universite-paris-saclay.fr} \\ \And
Lionel Mathelin \\ LISN-CNRS \\ Universit\'e Paris-Saclay \\ 91440, Orsay \\ France \\ \texttt{lionel.mathelin@lisn.upsaclay.fr}
}

\begin{document}

\maketitle

\begin{abstract}
The reliable prediction of the temporal behavior of complex systems is key in numerous scientific fields. This strong interest is however hindered by modeling issues: often, the governing equations describing the physics of the system under consideration are not accessible or, when known, their solution might require a computational time incompatible with the prediction time constraints. Not surprisingly, approximating complex systems in a generic functional format and informing it {ex--nihilo} from available observations has become common practice in the age of machine learning, as illustrated by the numerous successful examples based on deep neural networks. However, generalizability of the models, margins of guarantee and the impact of data are often overlooked or examined mainly by relying on prior knowledge of the physics. We tackle these issues from a different viewpoint, by adopting a curriculum learning strategy. In curriculum learning, the dataset is structured such that the training process starts from simple samples towards more complex ones in order to favor convergence and generalization. The concept has been developed and successfully applied in robotics and control of systems. Here, we apply this concept for the learning of complex dynamical systems in a systematic way. First, leveraging insights from the ergodic theory, we assess the amount of data sufficient for a-priori guaranteeing a faithful model of the physical system and thoroughly investigate the impact of the training set and its structure on the quality of long-term predictions. Based on that, we consider entropy as a metric of complexity of the dataset; we show how an informed design of the training set based on the analysis of the entropy significantly improves the resulting models in terms of generalizability, and provide insights on the amount and the choice of data required for an effective data-driven modeling.
\end{abstract}


\section{Introduction}\label{Sec_Intro} In the present era of mathematization of Nature, the need for modeling is ubiquitous. Focusing on engineering as an example, models are invaluable for predicting quantities of interest, designing and optimizing physical systems, understanding their behavior in an environment, allowing to derive control strategies. Physical models have historically been developed from a combination of observational data and first principles, depending on the amount and type of prior knowledge. This approach is still followed today where the mathematical structure of a model is typically provided and observational data are used to inform it, \textit{i.e.} learn the coefficients associated with the different terms of the retained structure. Countless variants exist but they generally follow this philosophy. Nonetheless, in many relevant cases, the physics of the phenomena is very complex and the interplay of observation with physical intuition is not sufficient to sort out a formalized model, let alone more rigorous general laws. In these cases, it is tempting to resort to a pure data-driven approach, in which some model is built up entirely from data. 
The data-driven approach received some attention after the work of Takens~\cite{TakensEmbedding, Eckmann_Ruelle_85, Sauer1991}, that provided the theoretical backbone to predictive models based on time-series analysis of low-dimensional systems~\cite{kantz2004nonlinear}. These approaches found new motivation following the significant progresses in the field of Machine Learning (ML), fostered by the massive augmentation of computer-based activities~\cite{goodfellow2016deep}. Even if a clear framework is still missing, the use of such algorithms to build models from data has been introduced in the last decades \citep[see \textit{e.g.}][]{voss1998identification, schmidt2009distilling} and, thanks to the recent blooming of data-driven techniques, has been considered in the modeling a variety of different physical phenomena \citep[see \textit{e.g.}][]{quade2018sparse, de2019unsupervised, brunton2020machine}. However, it is important to stress that the introduction of deep learning approaches has not necessarily changed the global picture of modeling. The potential versatility of highly expressive neural networks is often associated with a large amount of necessary observational data to suitably inform the associated coefficients of the model. Yet, not every situation enjoys a massive amount of observational data and can afford to inform a widely generic model. In practice, it is more likely that data are scarce due the cost and/or technical difficulty in obtaining them and complex high-dimensional systems may need a quantity of data significantly exceeding the observing capability,~\cite{boffetta2002predictability}. In that sense, the lack of data for inferring physical models might have direct consequences on the relevance of these highly expressive models. A strategy to alleviate this limitation is to simplify the postulated structure of the model so that it is less data hungry, with simpler models here understood as involving fewer parameters to be informed from the available data, hence a weaker constraint on the amount of training samples. Another strategy is to leverage prior knowledge one has on the physical system under consideration to guide the structure and the learning process, hence saving on data. For instance, known symmetries can be enforced by design in the structure or in the representation format of the input data, effectively reducing the need for data. A similar idea applies to invariances one may know the system obeys (conserved quantities such as energy, translational or rotational invariances, etc.) or properties such as stability in a given sense. This physics-aware machine learning approach has now become popular and is the motivation of numerous recent efforts \citep[see][among many]{Raissi_etal_2017, ZHU201956, Kashinath_etal_21, VonRueden_2021}. In that sense, the modeling of chaotic dynamical systems is a natural play-field~\cite{pathak2017using, borra2020effective}. Remarkably, previous works seem to indicate an intriguing capability to give accurate forecasting of chaotic systems, even relatively high-dimensional~\cite{pathak2018model}, apparently improving over  deterministic embedding techniques~\cite{kantz2004nonlinear}. These results therefore deserve to be thoroughly analyzed.
In the present work, we aim to shed some light on the challenges in data-driven learning dynamical systems and to explore the idea of leveraging prior information, here in the form of known organization of the dynamics, to inform the process of collecting relevant observations. Specifically, we adopt a curriculum learning strategy for dynamical systems from first principles, and assess it with respect to common sampling strategies. Curriculum learning relies on the idea that machines, like humans, learn best if first exposed to ``easy'' situations prior to dealing with ``more complex'' ones when well into the learning process. A possible rationale is that curriculum learning can be interpreted as an adaptive regularization method for optimization of a non-convex objective function. By progressively introducing more complex samples, the associated cost landscape evolves from near-convex to a more complex function exhibiting many more local extrema. Sorting training examples in this way may hence favor the optimization process toward a better extremum. This concept has been mostly developed in robotics and control of systems, \cite{Sanger1994NeuralNL}, and was formalized more recently in the context of machine learning, \cite{Bengio_etal_2009}. While it still lacks a firm theoretical ground despite some efforts \citep{ICML-2018-WeinshallCA}, curriculum learning has been widely illustrated and successfully employed in a variety of applications, ranging from robotics to video games, from computer vision to natural language processing and reinforcement learning, \cite{Narvekar_etal_20, Soviany_etal_22}. 
What makes a situation, or training sample in the context of machine learning, ``easy'' or ``complex'' is case-specific. For instance, in classification problems, the complexity of a particular example presented to the machine may be associated with the distance to the decision surface (e.g., SVM) or with the amount of noise in the data. Here, we illustrate the impact of curriculum learning based sampling strategies, by considering the celebrated Lorenz'63 system. While deterministic and low dimensional, the Lorenz system features many challenges encountered in more complex configurations, such as sensitivity to initial conditions and an attractor dimension $\dAttrac$ significantly lower than the ambient dimension $\Dambient$, here $\dAttrac \simeq 2.06 < \Dambient = 3$. It is thus somehow a minimally complex unit and justifies its relevance for illustrating some challenges in learning dynamical systems. We associate the complexity of a training element with its information entropy, in the sense of Shannon, and recognize that the solution of dynamical systems in the vicinity of an unstable fixed point is well described by a linear model and exhibits a low entropy. This observation is key and allows a principled sampling strategy in collecting observations of increasing complexity from the system at hand as it evolves from an unstable fixed point. We explore and compare different strategies for sampling a dynamical system in the aim of learning its behavior and predicting its state over time. These strategies only differ by the region in the phase space where observational samples are taken from, every other aspects being identical (sampling frequency, number of samples, observed quantity, etc.). These different configurations are chosen to be typical of situations ranging from no prior information, besides an estimated upper bound of the dimension of the dynamical system up to approximate knowledge of some invariants of the system. We carry out a comprehensive statistical analysis in order to systematically test the impact that different sampling strategies have on the quality of the resulting models in terms of prediction abilities and generalizability. 
From an algorithmic point of view, recurrent structures such as auto-regressive models or recurrent neural networks are popular choices in such situations. In the present study, we focus on the LSTM (Long-Short Term Memory) architecture~\cite{hochreiter1997long} for its widespread use, including applications in prediction of dynamical systems behavior~\cite{gers2002applying}. Moreover, it has been recently shown that the performance given by different architectures including reservoir computing is similar~\cite{vlachas2020backpropagation}, although for low dimensional systems like Lorenz'63, LSTMs appear to give the best results. Given this choice, we also provide best-practice guidance concerning the possible impact of initial conditions and the memory effects on the results of supervised learning models, beside analyzing the possibility of using less data while preserving generalizability and quality based on curriculum learning strategies.
The paper is organized as follows. General principles in learning a system from data are briefly discussed in Sec.~\ref{Sec_learning} to set up the stage and define notations. The general class of models we consider in this work is presented in Sec.~\ref{Sec_LSTM}. The specific configuration we retain for the present discussion is the Lorenz'63 introduced in Sec.~\ref{Sec_Lorenz_strats}, together with the different sampling strategies. The results are gathered and presented in Sec.~\ref{Sec_results} before the discussion in Sec.~\ref{Sec_discussion} closes the paper.


\section{Theoretical background}\label{Sec_learning}

\subsection{Learning dynamical systems}\label{Sec_learning_dyn_sys} The present work lies in the framework of learning a model to approximate the time evolution of a state of the system at hand from observations $\bobs(t) \in \R^{\nobs}$. To do so, one needs a representation for the state and a dynamical model for describing its evolution in time. A representation of the state is given by a function $\obstostate$ mapping observations $\bobs_\timedom(t)$ over a measurement time horizon $t \in \timedom$, to a set of coordinates $\bv \in \R^{\nv}$. Several mappings have been proposed in the literature, including delays by time intervals $\Delta t$, $\bv_i=\bobs(t-i\Delta t)$ for $i=0,1,\ldots$, and time derivatives of varying orders $i$, $\bv_i=\bobs^{(i)}$, \cite{Crutchfield1987, Gibson1992}. To obtain the size of the coordinate space, we assume that the state of the system $\bu$ lies on an attractor $\attrac$ of dimension $\dAttrac < d$ that is a boundaryless bounded, smooth sub-manifold of $\R^{d}$. In this case, the Whitney embedding theorem~\cite{WhitneyEmbedding} shows that if the number of coordinates $\nv$ is such that $\nv > 2 \, \dAttrac$ then the dynamics of the state can be entirely captured by the new coordinate system. The Takens embedding theorem~\cite{TakensEmbedding, Sauer1991} provides a way to make this result practical by asserting that one can take this number of time delays. Recent results have extended this theorem by providing guaranteed recovery properties based on the measurement operator, type of attractor, and other parameters of the problem, \cite{Eftekhari2018}.
In this work, we assume no prior knowledge of the governing equations of the model and rely on a purely data-driven approach. The dynamical model for describing the time evolution of $\bv(t) \in \R^{\nv}$ can be formulated in various ways, including feed-forward neural networks, owing to their high expressiveness. As an alternative, for instance in case an upper bound of the attractor dimension is unknown due to process noise or measurement noise, a recurrent model $\recmodel$ can be employed for describing the dynamics of $\bvpred$. The model is parameterized by $\btheta \in \R^{\ntheta}$ and writes
\be
\bvpreddot(t) = \recmodel\left(\bvpred\right),
\ee
where a dotted quantity denotes its time-derivative. Learning a model for the system under consideration takes the form of a classical supervised learning problem based on the misfit between observations $\bobs_\timedom \equiv \bobs\left(\bu\left(t \in \timedom\right)\right)$ collected through a measurement operator $\measope: \bu \mapsto \bobs$ over a time domain $\timedom = \left[t_0, t_T\right]$ and predictions, in the sense of some norm $\normLM{\cdot}{\timedom}$. It typically results in an optimization problem in terms of the model parameters, possibly including a regularization term $\regul$:
\bea
\btheta \in \arginf_{\dumtheta \in \R^{\ntheta}} \normsqLM{\bobs(t) - \measope\left(\bvpred\left(t;\dumtheta\right)\right)}{\timedom} + \regul\left(\dumtheta\right), \nonumber \\ \mathrm{s.t.} \qquad \bvpreddot\left(t;\dumtheta\right) = \recmodeldum\left(\bvpred\right).
\label{eq_loss_optim}
\eea

\subsection{Collecting data}\label{Sec_collect_data} The main issue impacting the predictions based on observations is the quantity of data actually needed to approximate the dynamics in a given sense. The issue is related to the ergodic theory of dynamical systems, whose founding idea is that the long-time statistical properties of a deterministic system can be equivalently described in terms of the invariant (time-independent) probability, $\mu$, such that $\mu(S)$ is the probability of finding the system in any specified region $S$ of its phase space. If the trajectories of a $\dbu$-dimensional ergodic system evolve in a bounded phase space $\phasespace \subset \boldsymbol{R}^{\dbu}$, the Poincar\'e recurrence theorem \cite{poincare1899methodes} ensures that analogues exist as it proves that the trajectories exiting from a generic set $S\in \phasespace$ will return to such set $S$ infinitely many times. The theorem holds for any points in $S$, almost surely. It was originally formulated for Hamiltonian systems, but it can be straightforwardly extended to dissipative ergodic systems provided initial conditions are chosen on the attractor. In $\dbu$-dimensional dissipative systems, the attractor $\attrac$ typically has a dimension $\dAttrac < \dbu$. The probability $\mu \left(B^{\dbu}_{\bx}(\varepsilon) \right)$ of finding points on the attractor which are in the $\Dambient$-dimensional ball of radius $\varepsilon$ ( with resolution $\varepsilon \ll 1$) around $\bx \in \attrac \subset \R^\Dambient$:
\be
\label{Eq_frac-dim}
\mu \left(B^{\dbu}_{\bx}(\varepsilon) \right) = \int_{B^{\dbu}_{\bx}(\varepsilon)} \mathrm{d}\mu(\widetilde{{\mathbf x}}) \sim \varepsilon^{\dAttrac}.
\ee
When $\dAttrac$ is not an integer, the attractor and its probability measure are \emph{fractal}. The attractor is typically multifractal~\cite{paladin1987anomalous}, but we can assume without lack of generality that it is homogeneous. Given that the Poincar\'e theorem proves the existence of good analogues, the issue is shifted to finding out how much time is needed for a Poincar\'e cycle to end. This key problem was considered by Smolouchoski and resolved by Kac \cite{kac1959probability}, who showed that for an ergodic system, given a set $S \equiv B^{\dbu}_{\bx^\star}(\varepsilon)$, the mean recurrence time $\langle \tau_S(\bx^\star) \rangle$ relative to $S$ is inversely proportional to the measure of the set $S$, \textit{i.e.},
\be
\label{Eq_kac1}
\langle \tau_{S}(\bx^\star) \rangle \propto \frac{1}{\mu(S)} \, \sim \varepsilon^{-\dAttrac},
\ee
where the average $\langle \cdot \rangle$ is computed over all the points $\bx\in S$ according to the invariant measure. Therefore, if we require $\varepsilon$ to be small and if $\dAttrac$ is large, the average recurrence time becomes huge, a symptom of the curse of dimensionality in data-driven methods.  
    
\begin{figure}[t]
\centering
\includegraphics[width=0.5\textwidth]{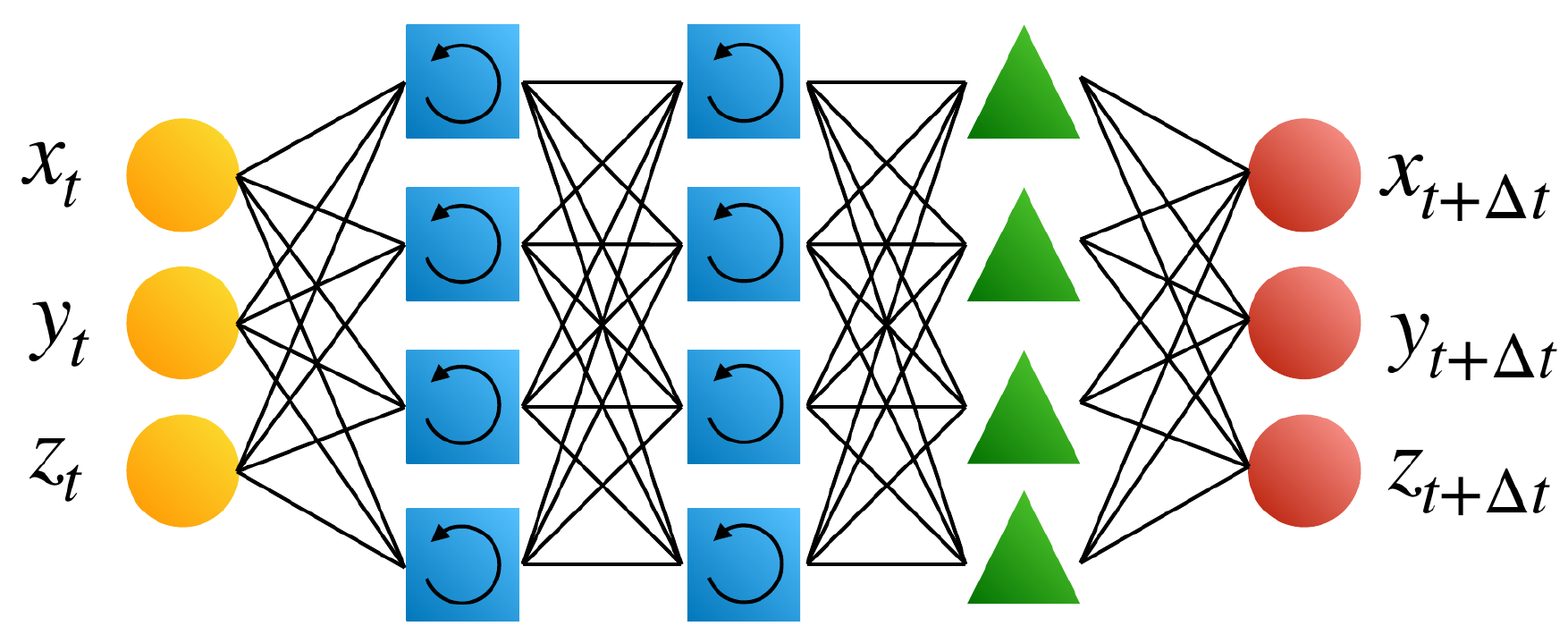}
\caption{Sketch of the retained architecture. Each hidden layer is made of LSTM cells, while the last layer is fully connected.}
\label{Fig_sketch_LSTM}
\end{figure}

\section{Machine-learning approach}\label{Sec_LSTM}

\subsection{Architecture}\label{Sec_archi} As discussed above, a set of $\Dambient$-dimensional observations retained to express a state of the system under consideration might not be a state vector, in the sense that it does not necessarily contain enough information to uniquely define its future evolution as an autonomous deterministic dynamical system. Even though the system were Markovian, it might not be so in the retained representation and may implicitly involve latent variables. This is typically the case in situations of partial observability or when measurements are affected by noise. As discussed in Sec.~\ref{Sec_learning}, under conditions rigorously studied in the embedology line of research, \cite{Eckmann_Ruelle_85, Sauer1991}, latent state variables can be substituted with observations over a recent past, that is, temporal information is traded for unavailable high-dimensional phase space information.
In the present work, we consider the LSTM (Long-Short Term Memory) architecture. Compared with standard multi-layer perceptrons, LSTMs additionally feature internal variables $\blatent$ carried-over through the recurrence and effectively acting as a ``memory''. Training an LSTM then allows to derive a discrete-in-time model predicting the future ``state'' $\bvpred_{k+1} \equiv \bvpred\left(t_k + \Delta t\right)$ from the current state and memory:
\be
\bvpred_{k+1} = \recmodel\left(\bvpred_k; \blatent_k\right).
\label{Eq_recmodel}
\ee
The retained architecture comprises 2 LSTM layers of 50 neurons each and one fully connected layer, with \verb?tanh? activation functions in the LSTM cells and \verb?sigmoids? for the state output (see the sketch in Fig.~\ref{Fig_sketch_LSTM}). The present architecture was deemed as a good compromise between accuracy and computational expenses, after an analysis with comparable architectures with one and three layers has been performed. A 1-step ahead model is learned based on a whole trajectory, or a set of trajectories, in the $L^2$-sense. The training is performed using the ADAM optimizer and an adaptive learning rate from $0.01$ to $10^{-5}$, with a decay factor of $0.07$. The maximum number of epochs is limited to $1000$.

\subsection{Strategies for sampling}\label{Sec_sampling} Learning a model here consists in estimating the best set of parameters $\btheta$ in the sense of the optimization problem defined in Eq.~\ref{eq_loss_optim}. In the context of neural networks, this acts as the loss function used for training. The regularization term may promote several properties of the optimal solution $\btheta$, such as the popular choice of low magnitude (ridge regression), $\regul\left(\btheta\right) \sim \normLM{\btheta}{2}$, or sparsity, $\regul\left(\btheta\right) \sim \normLM{\btheta}{1}$, possibly mimicked with a suitable dropout strategy.
We adopt a discrete-in-time viewpoint, in a formulation consistent with the discrete model in Eq.~\ref{Eq_recmodel}. The loss function $\loss$ associated with the training problem then writes
\be
\loss\left(\btheta\right) = \sum_{k=1}^K{\normLM{\bobs_k - \measope\left(\bvpred_k\left(\btheta \right)\right)}{2}^2} + \regul\left(\btheta\right),
\label{Eq_loss}
\ee
with $K$ such that $T = K \, \Delta t$ and the 1-step ahead dynamical model $\displaystyle \bvpred_{k+1} = \recmodel\left(\bvpred_k; \blatent_k\right)$.
The loss is here simply defined in terms of the Euclidean distance but many alternative definitions could be employed. For instance, optimal transport-based metrics, such as Dynamic Time Warping (DTW), can prove useful when observations come at a poorly known or controlled time pace. While the way the prediction from the model is compared with observational data is pivotal, the focus in this work is instead on the relevance of the training data. The present findings are believed to hold, disregarding the definition of the loss. A standard $L^2$ distance, implicitly assuming regularly sampled measurements, is hence retained.
The measurement operator is the identity, $\measope \equiv I$, so that the measurements are associated with the state of the system, $\bobs_k \equiv \bu_k$, $\forall \, k$. No noise affects the observations, so that one is in a full observability situation. In this context, the observed system is Markovian in the representation space of the observations. However, a recurrent architecture is here considered as a general structure for learning, relevant for a wide class of situations. It further has the benefit of improving upon the robustness of the learning by virtue of some degree of redundant representation.
Without loss of generality, we disregard the regularization term $\regul\left(\btheta\right)$ in the definition of the loss function which finally writes
\bea
\loss\left(\btheta\right) = \sum_{k=1}^K{\normLM{\bobs_k - \bvpred_k\left(\btheta \right)}{2}^2},
\nonumber \\
\mathrm{s.t.} \qquad \bvpred_{k+1} = \recmodel\left(\bvpred_k\right), \quad \bvpred_0 = \bobs_0.
\eea
While this context is favorable for learning (low-dimensional deterministic system, noiseless observations, full observability, etc.), it will be seen below that the measurement strategy still significantly affects the learning performance.

\subsection{Curriculum learning, metrics and complexity of the datasets}\label{Sec_complexity} As mentioned in the Introduction, we aim at verifying the extent to which a measurement strategy inspired by the curriculum learning principle can be applied to data-driven learning of dynamical systems and how this impacts the generalizability and quality of the resulting models. The rationale is the adaptive regularization of the objective function when moving from ``simpler'' cost landscapes associated with simpler data towards more complex ones. 
From this perspective, any indicator of complexity can be used; here, we consider entropy as a measure and compute it based on singular value decomposition (SVD) \cite{varshavsky2006novel}. Given an embedding for the measurements $\bobs$ based on time delaying
\bea
\bobs = \left[s(k),\,s(k+\tau),\,\dots,\,s(k+(n_e-1)\tau)\right],
\eea
with $k$ an integer and $\tau$ the time delays, the matrix $\bObs$ is defined as
\bea
\bObs = \left[\bobs(0),\bobs(\tau),\,\bobs(2\tau),\,\dots,\,\bobs(N-(n_e-1)\tau)\right],
\eea
with $n_e$ indicating the number of $\tau$. The SVD is performed on the matrix and $M$ singular values $\left\{\sigma_i^{\rm SVD}\right\}_i$ are obtained such that the entropy $H$ is defined as 
\bea
H_{SVD} = -\sum_{i=1}^M \overline{\sigma}^{\rm SVD}_i \log_2 (\overline{\sigma}^{\rm SVD}_i),\label{eq-svd}
\eea
with $\overline{\sigma}^{\rm SVD}_i = \sigma^{\rm SVD}_i / \sum_{j=1}^M \sigma^{\rm SVD}_j$. The SVD entropy provides an estimation of the number of eigenvectors that are needed for an adequate representation of the dataset. In that sense, it measures the dimensionality of the data. With this metric, we will assess in the next section the complexity characterizing each of the training datasets that will be compared. 

The second metric introduced in the article for the evaluation of the models is the correlation dimension, computed using the algorithm by Grassberger \& Procaccia \cite{grassberger1983characterization}. Considering a ball of radius $\varepsilon$, the probability of finding a point scales as $C(\varepsilon) \sim \varepsilon^{d_2}$, where $d_2$ is an indication of the dimensionality. Using the Euclidian norm, the probability $C\left(\varepsilon\right)$ can be efficiently computed as the correlation sum
\begin{equation}
C\left(\varepsilon\right)= \lim_{N\rightarrow\infty}\dfrac{1}{N \left(N-1\right)}
\sum_{\substack{i,j=1 \\ i \neq j}}^N
\Phi\left(\varepsilon-\|s_i-s_j\|\right),\label{eq:corred2}
\end{equation}
with $\Phi$ the Heaviside step function, $\bobs$ the components of the embedded variable ranging $i,j=1,2\dots N$, with a fixed time increment; $N$ is the number of pairs. Due to the divergence of trajectories, most pairs $(s_i,s_j)$ with $i\neq j$ are usually uncorrelated. Here the correlation dimension ${d_2}$ is considered for the information provided on the fractal dimension of the attractor as well as the connections with the intrinsic dimension, a metric often found in machine learning \cite{camastra2016intrinsic}. Alternatively, the 
Lyapunov exponents can also be adopted as criteria.

\begin{figure}
\centering
\begin{subfigure}[t]{0.475\textwidth}
\centering
\includegraphics[width=\textwidth]{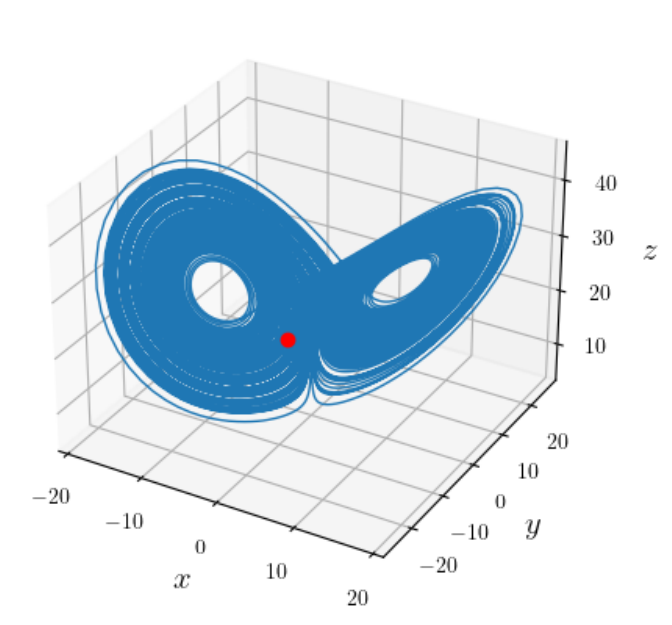}
\caption{Ergodic datasets: $\tsetergo$ and $\tsetsplit$.}\label{Fig_trajs_ergo}
\end{subfigure}
\hfill
\begin{subfigure}[t]{0.475\textwidth}
\centering
\includegraphics[width=\textwidth]{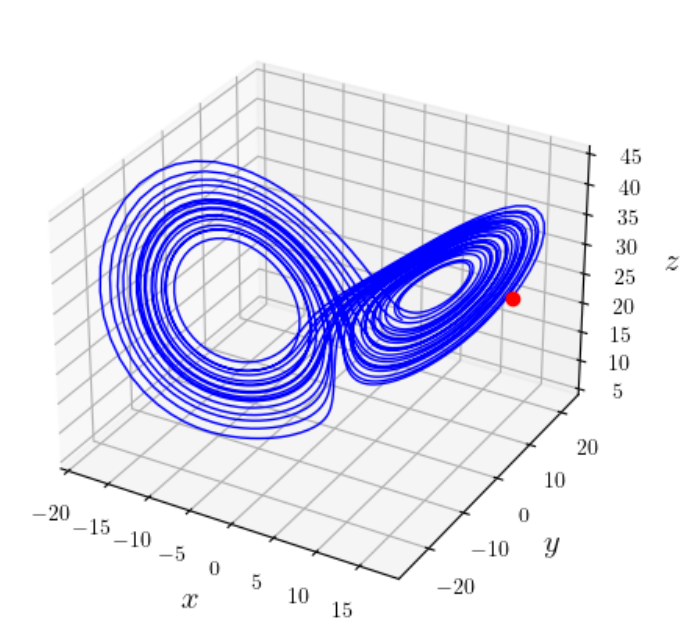}
\caption{Random sampled dataset: $\tsetrand$.}\label{Fig_trajs_ergo}
\end{subfigure}
\hfill
\begin{subfigure}[t]{0.475\textwidth}
\centering
\includegraphics[width=\textwidth]{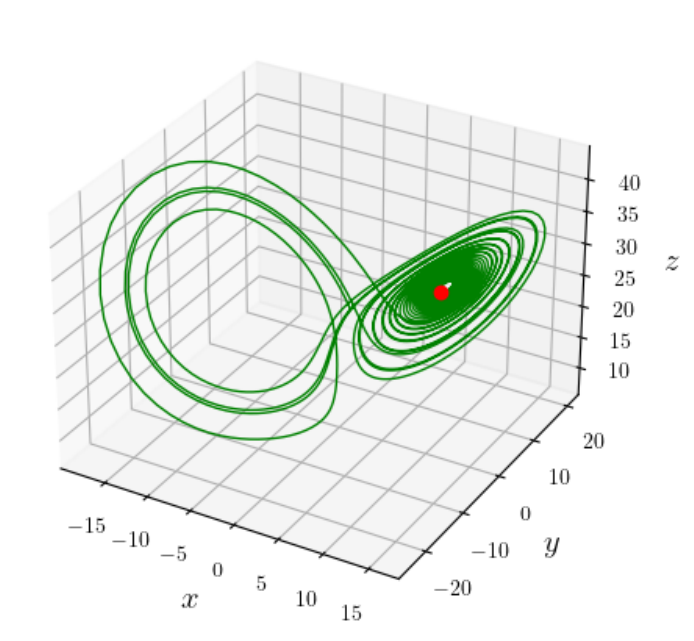}
\caption{Structured dataset: $\tsetFP$.}\label{Fig_trajs_ergo}
\end{subfigure}
\caption{The figure qualitatively illustrates the different training sets used in this investigation. Each ergodic dataset ($\tsetergo$) is composed by one long trajectory $(a)$. The datasets with trajectories emanating from the fixed point, exploiting the structure of the attractor ($\tsetFP$) and the datasets randomly sapled ($\tsetrand$) are composed by an ensemble of $9$ trajectories, of which one example for each is shown in $(b)$ and $(c)$, respectively (see also Sec.~\ref{Sec_setup}).}\label{Fig_sampling_trajs}
\end{figure}

\section{Learning the Lorenz'63 system}\label{Sec_Lorenz_strats} 

\subsection{Lorenz'63 system} To illustrate the different learning strategies, we consider the celebrated Lorenz'63 system~\cite{lorenz1963deterministic}, originally introduced to mimic the thermal convection in the atmosphere. It consists of three coupled nonlinear ordinary differential equations (ODEs):
\bea
\begin{cases}
\dot{x} = \sigma \left(y - x\right), \nonumber \\
\dot{y} = \rho x - y - x z, \\ 
\dot{z} = x y -\beta z,\nonumber
\label{eq:lorenz}
\end{cases}
\eea
with $\left\{\sigma, \rho, \beta\right\} \in \R^+.$ The system of ODEs is integrated in time with a $4^{th}$ order Runge-Kutta scheme and a time step $\Delta t = 0.01$. Constants $\sigma, \rho, \beta$ are positive parameters, related to dimensionless scales such as the Prandtl and the Rayleigh numbers. For $\rho > \sigma \left(\sigma+\beta+3\right) \slash \left(\sigma-\beta-1\right)$ and $\sigma>\left(\beta+1\right)$, the solution $\bu := \left(x,y,z\right)$ exhibits a chaotic behavior. It is important for the following to note that the system features three unstable fixed points $\fp$, associated with a vanishing time-derivative:
\bea
\fpp & = & \left(\sqrt{\beta(\rho-1)},\sqrt{\beta(\rho-1)},\rho-1\right), \nonumber \\ 
\fpm & = & \left(-\sqrt{\beta(\rho-1)},-\sqrt{\beta(\rho-1)},\rho-1\right), \nonumber \\
\fpzero & = & \left(0,0,0\right). \nonumber
\eea
In the following we shall consider the parameters fixed to $\beta=8/3, \sigma=10, \rho=28$ as in the original work, such that the solution is in the chaotic regime. In this regime, the Kaplan-Yorke dimension of the attractor is $\dAttrac \simeq 2.06$. 

\subsection{Set-up of the learning method}\label{Sec_setup} We turn our attention to the training set $\trainset = \left\{\bobs_k\right\}_{k=1}^K$ for learning the Lorenz'63 system. It is a crucial ingredient of the learning process and largely conditions the quality of the resulting model. We consider four different training sets ($\tsetergo$, $\tsetsplit$, $\tsetrand$, and $\tsetFP$ ) of identical settings (sampling frequency, number of samples, measurement operator, etc.) but following different sampling strategies.
\subsubsection{Ergodicity-compliant training sets, $\tsetergo$ and $\tsetsplit$}\label{Sec_Tergo} The $\tsetergo$ dataset is a long trajectory on the attractor. We consider it the reference case as it provides sufficient statistics for learning a reliable and accurate model. With $\varepsilon = 10^{-2}$ and the Kaplan-Yorke dimension $\dAttrac \simeq 2.06$ of the Lo-renz'63 for the retained settings $\left\{\sigma,\rho,\beta\right\}$, \cite{Kuznetsov_etal_2020}, the required number of samples resulting from a constant sampling period of $\Delta t = 0.01$ is about $27,000$ strobes in time. A second dataset, $\tsetsplit$, is derived from $\tsetergo$: it consists of one, long trajectory on the attractor, yet split in 9 chunks of equal size and reshuffled, such that each chunk is thus composed by $3000$ samples. As it will be clear when discussing the results (Sec.~\ref{Sec_results}), this dataset will allow us to investigate some effects of memory.

\subsubsection{Structured datasets: leveraging the fixed points, $\tsetFP$}\label{Sec_TFP} The training set $\tsetFP$ exploits prior knowledge on the structure of the system at hand, namely the fixed points. As mentioned in the Introduction, it is well known dynamical systems are organized around critical features, such as fixed points, and that invariant sets provide information on the behavior of the solution, \cite{Gilmore_etal_10}. Besides fixed points, these features include higher-dimensional invariant sets such as slow invariant manifolds. The sampling strategy here consists in acquiring observations from the system originating from its (unstable) fixed points, in contrast with other training sets obtained from the attractor. Specifically, three $3000$-sample trajectories are acquired from each of the three fixed points of the Lorenz'63 system. These trajectories originate from a small deviation from each fixed points along each of the $\Dambient=3$ directions given by the eigenvectors of the local Jacobian matrix. Note that if the Jacobian is not available, a set of orthonormalized deviations can also be used. The resulting training set then consists of $3 \times 3 \times 3000 = 27,000$ samples, consistent in size with the other training sets illustrated in Fig.~\ref{Fig_sampling_trajs}.

\begin{figure}
\centering
\includegraphics[width=.475\textwidth]{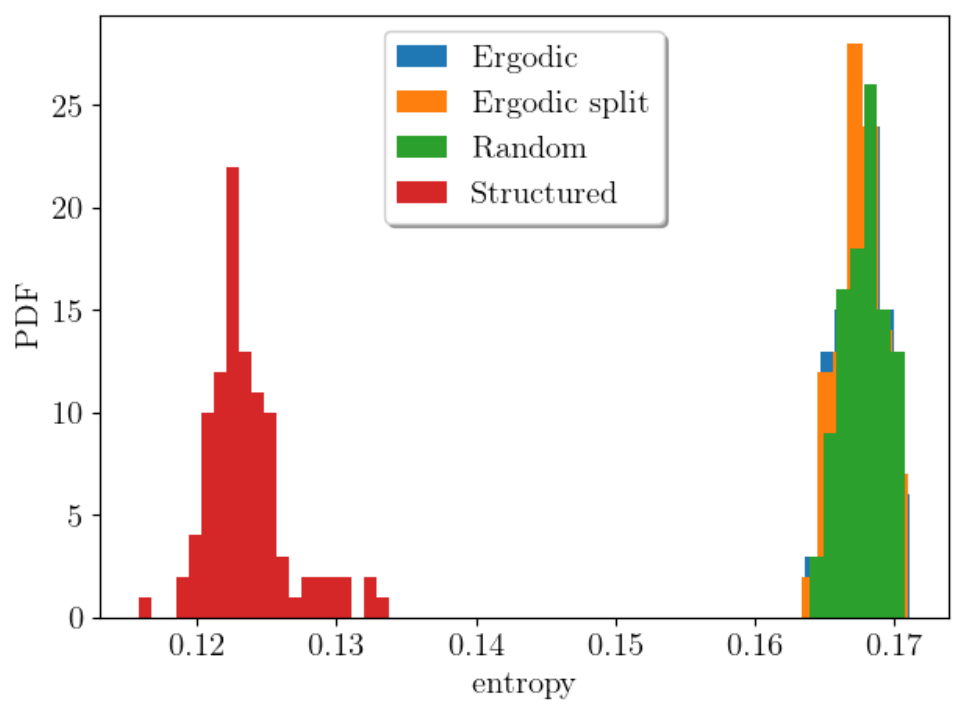}
\caption{Distribution of the SVD-entropy (Eq.~\ref{eq-svd}) associated with the four different datasets described in Sec.~\ref{Sec_setup}: ergodic ($\tsetergo$), ergodic-split ($\tsetsplit$), random sampled ($\tsetrand$) and structured ($\tsetFP$). Only the structured dataset shows substantially lower entropy, while the other datasets share the same values and distribution.}\label{Fig_traj_entro}
\end{figure}

\subsubsection{Random sampling, $\tsetrand$}\label{Sec_Tens} Finally, we consider the sampling of short trajectories on the attractor. Nine trajectories are considered, each originating from a state randomly chosen on the attractor. Each trajectory is $3000$ sample long as for the other strategies introduced above so that the total size of the training set is again $9 \times 3000 = 27,000$ samples. This training set is representative of the widespread situations where information about a system come in several independent glimpses, as resulting from several measurement campaigns. It is illustrated in Fig.~\ref{Fig_sampling_trajs}.

\begin{figure*}[t]
\centering
\includegraphics[width=1.05\textwidth]{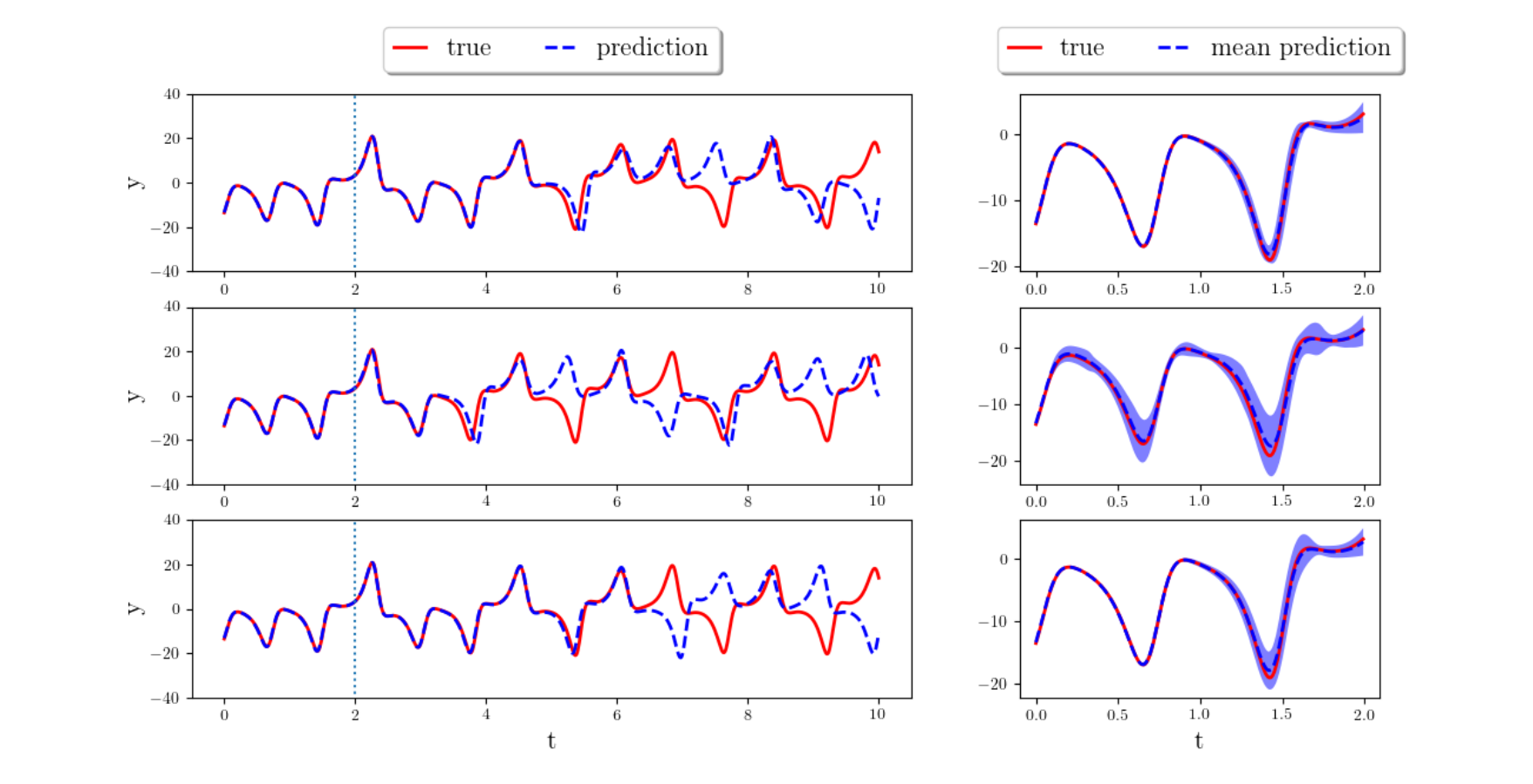}
\caption{Left column: Time-evolution of the $y$-component of the Lorenz'63 system, both from the true (\texttt{true}, solid red) and the learned model (\texttt{prediction}, dashed blue) from different training sets and initial conditions. Right column: zoom on the [0, 2] time units interval, with the mean prediction (dashed blue line) from an ensemble of 100 trained models, and the associated 1 standard deviation.  Top row: The prediction of the best (out of the ensemble set) model trained from $\tsetshort$ is shown. The initial condition is drawn from the training set. The prediction is seen to remain reliable up to about 4.3 Lyapunov times (5 time units) before drifting significantly from the truth. Middle row: when the same model is evaluated with an unknown initial condition (test set), the performance deteriorates and the accuracy is lost after less than 4 time units. When a unique ergodic-compliant trajectory $\tsetergo$ is used to train a model (bottom row), the prediction from an unknown initial condition improves significantly.}\label{Fig_tempevo}
\end{figure*}

\subsubsection{Entropy characterization}\label{Sec_EntDS} Using the definition of SVD entropy provided Eq.~(\ref{eq-svd}) with order $n_e=3$ and time-delay $\tau=1$, we characterize in the following each of the training sets. The SVD-entropy algorithm has been chosen after a comparison among different strategies for the entropy estimation. First, we considered convergence of the estimates over long trajectories emanating from different initial conditions, in order to evaluate the consistency of the results in ergodic datasets; second, a validation based on entropy estimators requiring a reduced number of data, namely the Approximate Entropy algorithm by \cite{pincus1991approximate}, has been performed. Good agreement was found between the different entropy estimators. From the physical viewpoint, the entropy for a time series is defined on the probability transition between the embedded states (unfolded through SVD).
In Fig.~\ref{Fig_traj_entro}, the entropy distributions are compared for each of the training sets: quite neatly, the dataset $\tsetFP$ is associated with the lowest entropy while the remaining datasets are all characterized by the same levels. From the physical viewpoint, this behavior is explainable by considering the higher predictability of the dynamics in the vicinity of the fixed points, where the behavior of the states close to them is well approximated by the linear limit. High predictability corresponds to less required information. As we will show in the following, this leads to an apparent contradiction: despite less information is contained in the structured datasets -- in a Shannon sense -- the possibility of training LSTM with well structured datasets characterized by increasing entropy in time allows to deploy a curriculum learning strategy.

\begin{figure}[t]
\centering
\includegraphics[width=.4925\textwidth]{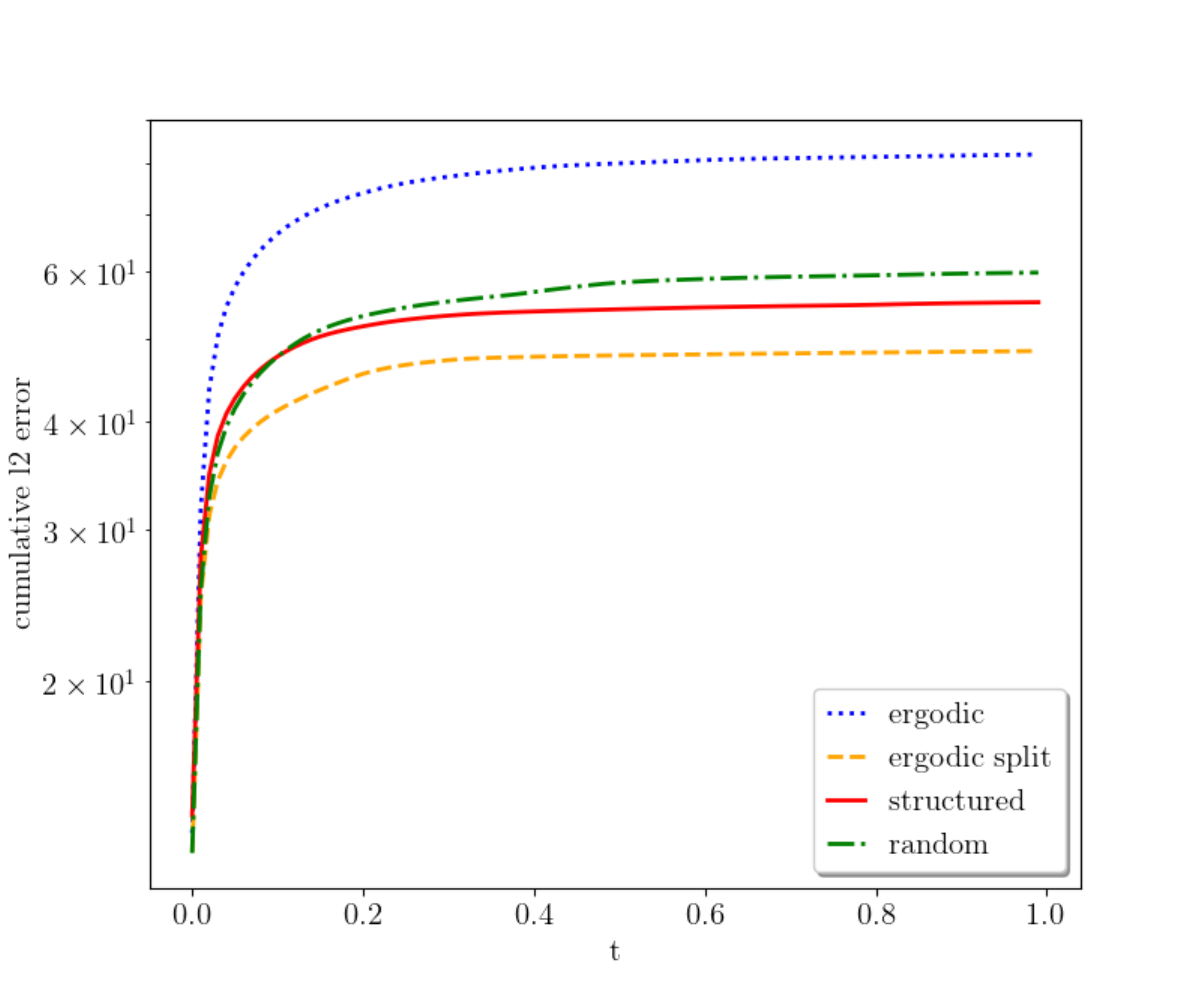}
\includegraphics[width=.4925\textwidth]{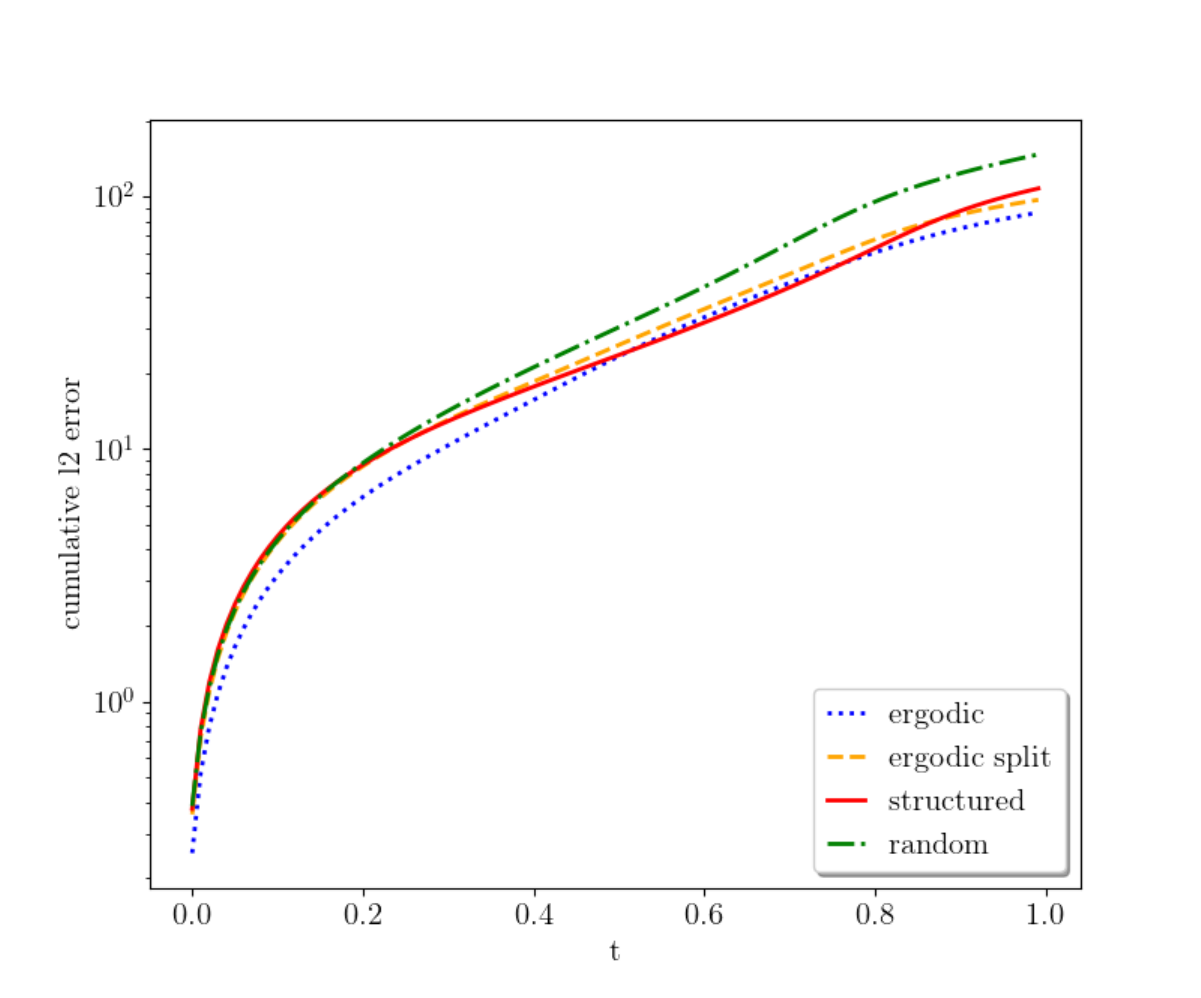}
\begin{picture}(0,0)
\put(-435,160){$(a)$}
\put(-202,160){$(b)$}
\end{picture}
\caption{Time evolution of the prediction analysed as cumulative $2$-norm error, using an average of $100$ models. A short windows of $1$ time unit is considered. In the inset $(a)$, the average predictions emanating from trajectories included in the training set are shown. In $(b)$, the error is shown when analysing the test sets: larger deviations are observed for the random sampled datasets.}
\label{Fig_tempevo2}
\end{figure}

\begin{figure*}
\centering
\includegraphics[width=0.9\textwidth]{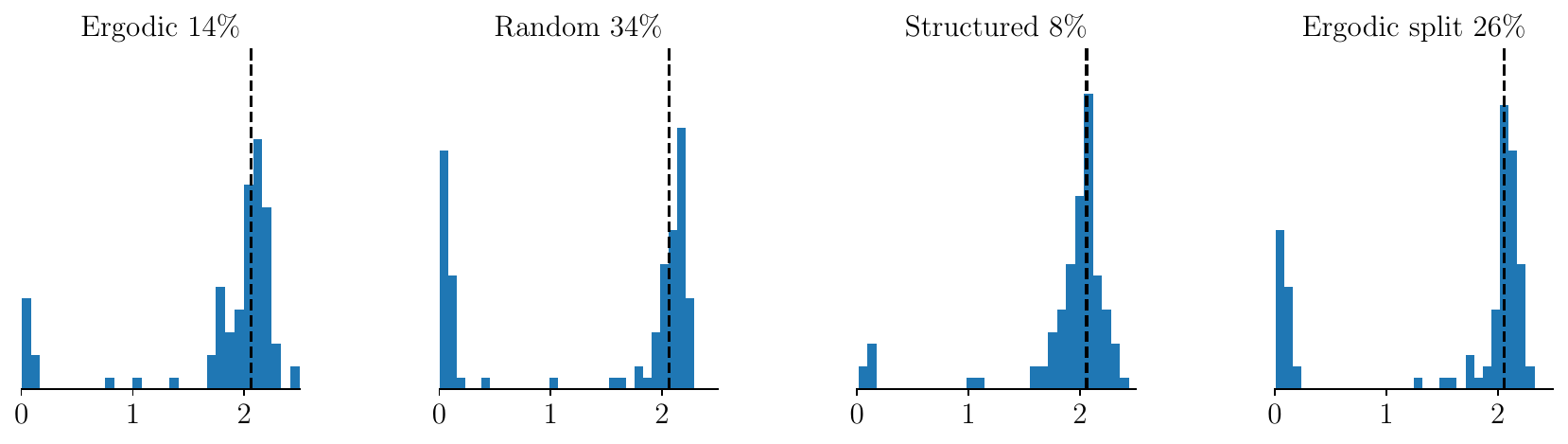}
\caption{The $d_2$ dimension is here adopted as a metric for assessing the prediction quality of the 100 models trained with each of the $4$ training strategies: ergodic ($\tsetergo$), random-sampled ($\tsetrand$), structured ($\tsetFP$), and ergodic-split ($\tsetsplit$); the resulting $d_2$ dimension is estimated by analyzing time predictions from trajectories not belonging to the training set. For each training set, the proportion of models associated with an estimated dimension differing by more than 25\% from the true dimension $\dAttrac=2.06$ is indicated in percent.}
\label{Fig_d2_0mem}
\end{figure*}

\section{Results} \label{Sec_results} We now assess the quality of the models learned from various observation datasets of the Lorenz'63 system.

\subsection{How much training information?} \label{Sec_Generalisation} The influence of the amount of information to train a model is first considered. Two models with the same architecture (cf. Sec.~\ref{Sec_LSTM}) are trained from two different datasets. These datasets consist of observations of the solution of the Lorenz system along the \emph{same} trajectory on the attractor, but of different length. One dataset, $\tsetergo$, comprises 27,000 samples while the second one $\tsetshort$ is its restriction to the first 3000 samples, so that $\tsetshort \subset \tsetergo$. Once trained, the resulting models, hereafter respectively termed $\recergo$ and $\recshort$, are evaluated in terms of their long-term prediction capability. Results are gathered on Fig.~\ref{Fig_tempevo}. The performance of the models is here evaluated when predicting from a known (\textit{i.e.}, contained in their training set) and unknown (not contained) initial condition, to assess the generalization capability of the models. For all cases, memory is initialized to zero.

\begin{figure}[t]
\centering
\begin{subfigure}[t]{0.49\textwidth}
\centering
\includegraphics[height=0.8\textwidth]{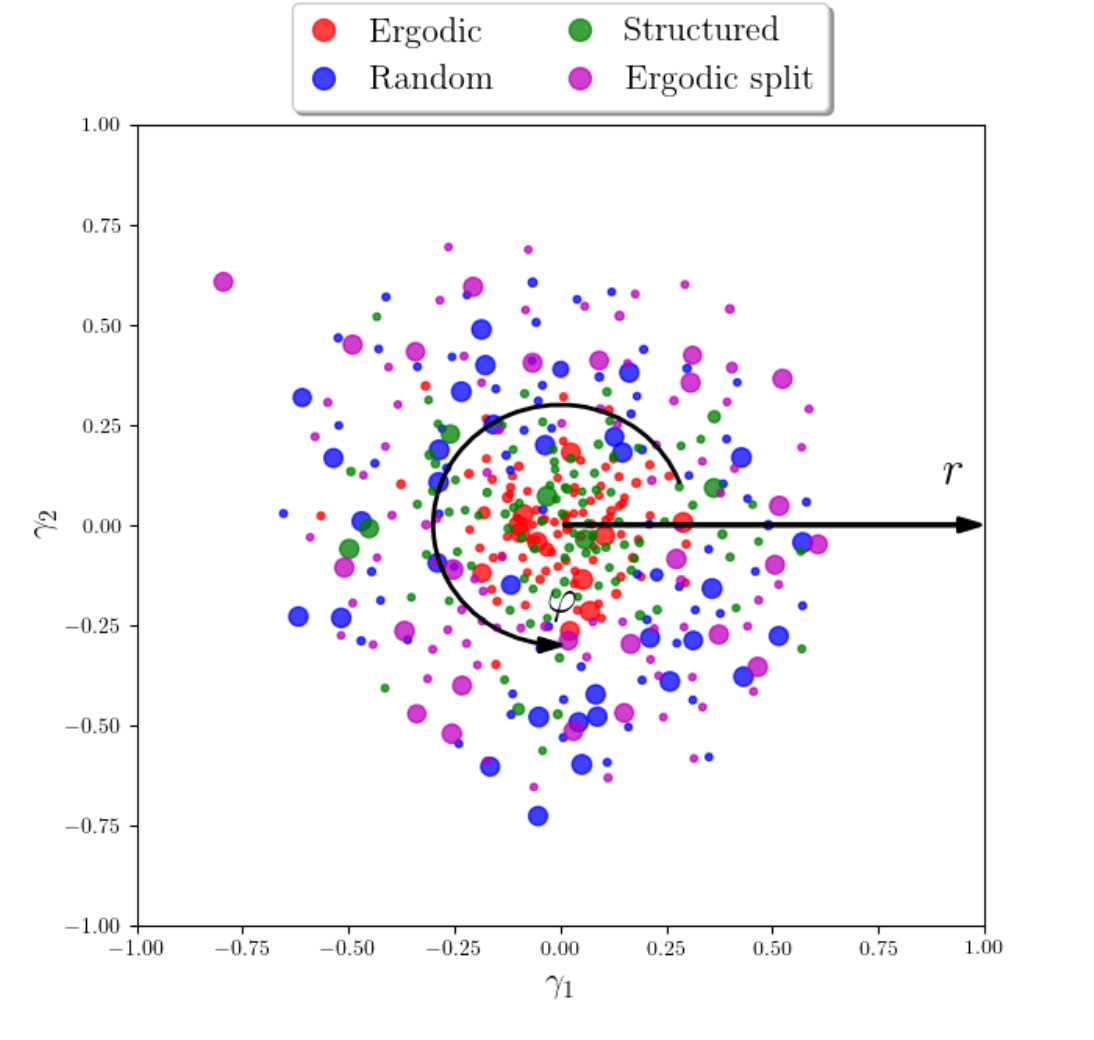}
\caption{t-distributed stochastic neighbor embedding (t-SNE) of the parameters of the trained models analyzed in Fig.~\ref{Fig_d2_0mem}. Each dot corresponds to a different model and its size visually indicates the accuracy: the larger the dot, the larger the error with respect to $\dAttrac$.}
\label{Fig_TSNE_0mem}
\end{subfigure}
\hfill
\begin{subfigure}[t]{0.49\textwidth}
\centering
\includegraphics[height=0.8\textwidth]{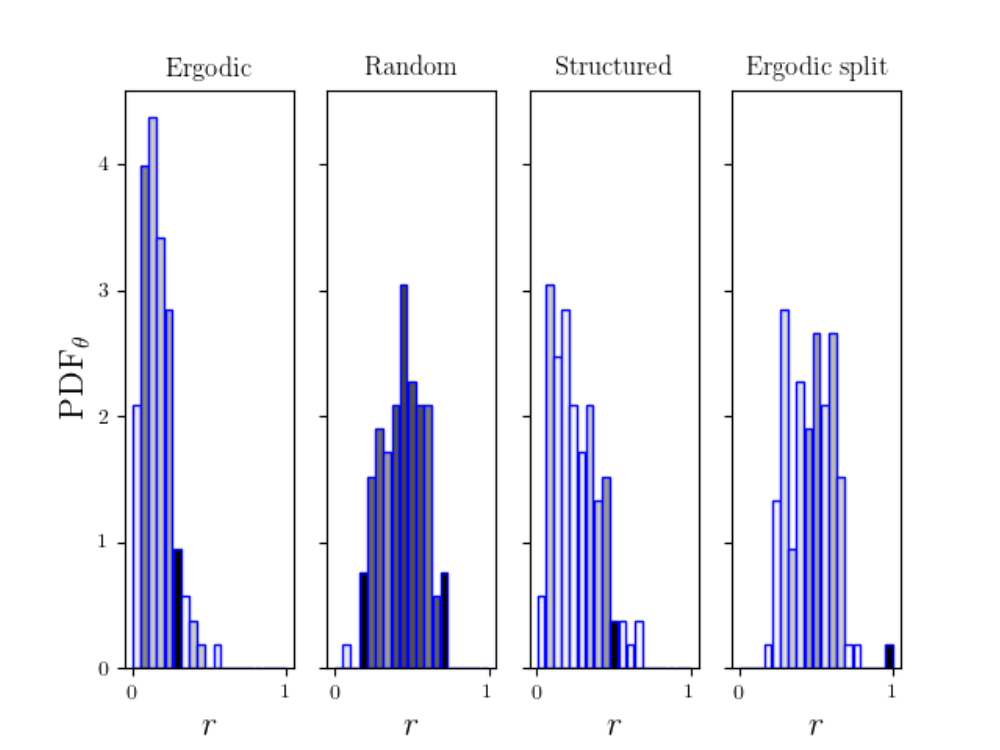}
\caption{Radial distribution of the t-SNE projected model parameters. $r$ indicates the distance from the barycenter of the cluster in Fig.~\ref{Fig_TSNE_0mem}, while the distribution of the models is reported on the vertical axis using the same number of bins. We order from light cyan to darker blue models performing progressively worse in terms of resulting $d_2$ accuracy; note that the level set of the blue gradient is normalized for each subplot, such that only a qualitative information on the distribution along $r$ is included in the graphs.}
\label{Fig_radTSNE_0mem}
\end{subfigure}
\caption{Illustration of the model parameters via a t-SNE representation. The LSTM memory is initialized to zero.}
\label{fig_Zero_mem}
\end{figure}

A known initial condition is first considered. As can be seen from the top row (left column) where it is plotted against the truth (red), the $\recshort$ model allows a reasonable quality of the prediction (plotted in blue) over a significant time horizon of about 6 Lyapunov times (7 time units). Since training the models is achieved via a stochastic procedure, \textit{e.g.} stochastic gradient descent and random initialization of the parameters, and the landscape of the loss function is not globally convex, the underlying optimization problem ends up in a local optimum and bears an aleatoric contribution. To strengthen the above conclusions to account for the statistical uncertainty of neural network modeling, we hence consider an ensemble of 100 identical independent models and evaluate their performance \cite{agarwal2021deep}; the final evaluation is thus based on an aggregate metric that allows to provide a robust assessment of the relative performance across different trainings based on the different datasets. The right column of Fig.~\ref{Fig_tempevo} shows the mean and standard deviation of the prediction of the ensemble of models for the same conditions as in the left column. From the top row, one may deduce that the prediction of the ensemble of models is very consistent and that the learning is robust.
With the same short training set $\tsetshort$, the trained model is now evaluated on an initial condition not contained in the training data, see Fig.~\ref{Fig_tempevo} middle row. Compared with the previous situation (top row), the performance is seen to deteriorate, both for a single model (left column) and in average (right). In particular, the time horizon before the prediction significantly differs from the truth is much shorter, indicating a poor generalizability properties. When more training data are used, $\tsetergo$, the resulting model remains accurate over a long time horizon even when predicting from an unknown initial condition, as seen from Fig.~\ref{Fig_tempevo} (bottom row). Again, this is a reliable finding as evidenced from the ensemble performance seen on the right column. 

In Fig.~\ref{Fig_tempevo2} a complementary analysis is shown, considering the average cumulative $2$-norm error over a short time period of $1$ time-unit. The averaging is performed considering $100$ models for each of the $4$ sampling strategies. The models are trained using random initialization of the memory. In the inset $(a)$, the prediction emanating from trajectories included in the training set is shown. It is possible to see that the error is bounded and constant, at least within the short time window we consider here; the ergodic case is characterized by higher error due to a larger number of strobes back--propagated during the training in each batch. In $(b)$, the error is shown for the testing sets. First -- coherently to Fig.~\ref{Fig_tempevo} -- the error increases while time increases; secondly, larger deviations are observed for the random dataset, while the cumulative error of the other cases is more comparable.

These results illustrate the fact that the training data, relied on to learn a model, have to be informative enough of the system under consideration to allow for generalizable models to be inferred. In particular, the superior performances shown in the top row of Fig.~\ref{Fig_tempevo} and the low error shown in Fig.~\ref{Fig_tempevo2}$(a)$, when the initial conditions are contained in the learning data-set appear to be due to overfitting.

\begin{figure*}
\centering
\includegraphics[width=0.9\textwidth]{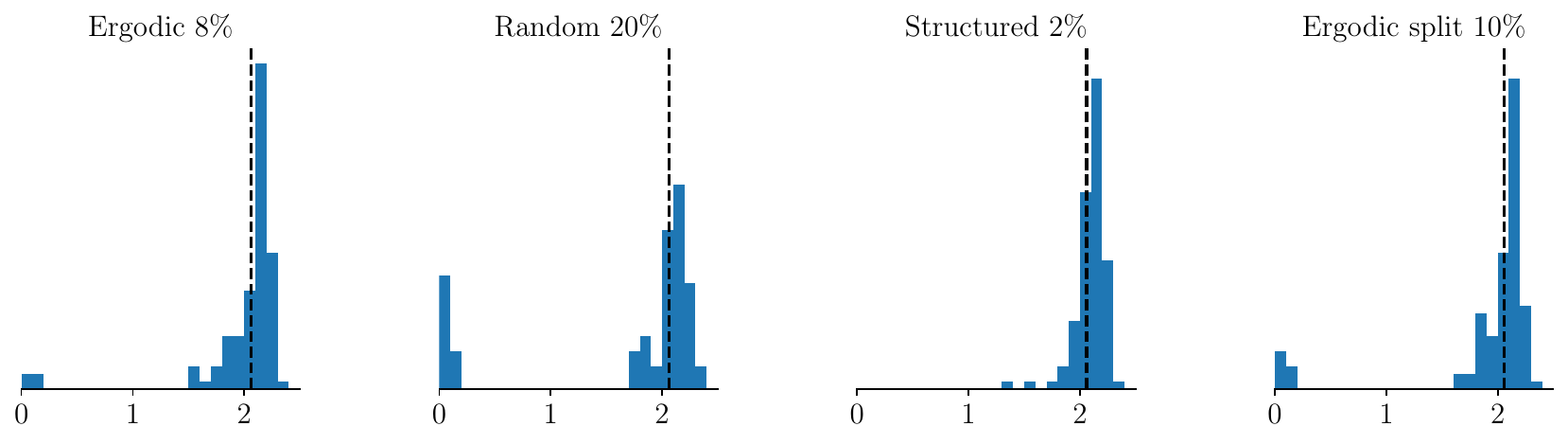}
\caption{The figure depicts the empirical distribution of the $d_2$ dimension associated with the predictions obtained for models trained by initializing the LSTM memory at random, using the four different learning strategies. For each training set, the proportion of models associated with an estimated dimension differing by more than 25\% from the true dimension $\dAttrac=2.06$ is indicated.}
\label{fig_rand_mem}
\end{figure*}

\subsection{Relevance of the data}\label{Sec_which_training} Besides the mere size of the training set, a critical aspect is its \emph{relevance} for informing a model in view of a given objective function. To shed light on this pivotal aspect, the four different training sets introduced in Sec.~\ref{Sec_sampling}, namely the ergodic ($\tsetergo$), the ergodic-split ($\tsetsplit$), the random-sampled ($\tsetrand$) and the structured based on the fixed points ($\tsetFP$), all of the same size of $27,000$ samples, and analyzed using the SVD entropy in Sec.~\ref{Sec_EntDS}, are now evaluated in terms of the quality of the associated resulting model. To avoid any bias due to the initialization of the parameters, an ensemble of $100$ models is trained for each situation and the comparison is made in terms of the ensemble performance \cite{agarwal2021deep}.
A first metric to quantify the quality of the resulting model is the $d_2$-dimension of the predicted model, computed by estimating the correlation sum in Eq.~\ref{eq:corred2}. This criterion is not associated with a given initial condition and is affected by the way the trained model consistently reproduces the fractal nature of the Lorenz'63 system. Results are gathered in Fig.~\ref{Fig_d2_0mem} in terms of a histogram of the estimated $d_2$ dimension for the ensemble of 100 models for each of the four training sets. For a more quantitative appreciation, a threshold is defined when the dimension resulting from the trained model differs by more than 25 \% from the truth $\dAttrac = 2.06$. 
As expected, the ergodicity-compliant training set $\tsetergo$ leads to a good performance, with most models associated with a $d_2$ dimension close to the true value and a few degenerate models such that $d_2 \approx 0$. Only 14 \% of the models are beyond the threshold. Models trained from $\tsetFP$ are also seen to achieve a similar level of performance. Again, almost all these models are associated with a rather good $d_2$ dimension, with very few degenerate models. Only 8\% of the models are beyond the threshold, an even better performance than models trained from $\tsetergo$.
In contrast, the $\tsetrand$ (random) and $\tsetsplit$ (split) dataset leads to a rather poor performance model, with a significant part of the models associated with $d_2 \approx 0$ and respectively 34 and 26 \% of the models beyond the threshold.
An observation is in order in this regard. The models with $d_2 \approx 0$ are all characterized by being stationary: the dynamics of the system stabilizes towards the fixed points after a short transient. On the other hand, only a limited number of models is characterized by $d_2\approx 1$; in this case the temporal dynamics settles on periodic orbits. Interestingly, in most of the stable models, the fixed points are correctly predicted, also for the random datasets where they are not included explicitly; however, their stability features are not correctly predicted, as already stressed.

It may be not quite surprising that models trained with $\tsetrand$ and $\tsetsplit$ are associated with a similar performance. Since the only difference between the two datasets is that the samples in $\tsetrand$ originate from a collection of 9 \emph{independent} segments of trajectory, randomly located on the attractor, while $\tsetsplit$ is made of 9 \emph{contiguous} segments. Yet, such behavior for the  $\tsetsplit$ dataset appears to be not trivial, because it contains exactly the same information of the $\tsetergo$ one about the phase space.
Further assessment of these observations can be made using a different technical tool, namely t-distributed stochastic neighbor embedding (t-SNE) of the parameters, shown in Fig.~\ref{Fig_TSNE_0mem}. It is essentially a dimensionality reduction technique convenient for assessing high-dimensional quantities~\cite{van2008visualizing}. The model parameters vectors are here mapped onto a 2-dimensional space where each model can hence be represented in terms of its corresponding coefficients $\gamma_1$ and $\gamma_2$. $\recergo$ and $\recFP$ are mostly clustered in the center of the points cloud, indicating that, while the models are different, they are however similar to each other in terms of parameter distribution. Instead, the models obtained from the random sampling strategy ($\tsetrand$) and the ones obtained stacking chunks from an ergodicity-compliant trajectory ($\tsetsplit$) are distributed in the outer region. LSTM parameters are initialized as independent realizations of a Gaussian random variable, whose t-SNE representation is typically visually close to a Gaussian as well since it penalizes deviation from a t-Student distribution. Recent works have shown that neural networks associated with parameters weakly evolving during training while yet able to significantly improve the loss function, are often associated with good generalization properties, \cite{NTKpaper}. In the present study, the t-SNE representation of models trained from the ergodicity-compliant and the structured datasets both appear close to a Gaussian distribution, a further indication that they might enjoy good generalization properties. In contrast, the t-SNE distribution of the models trained from $\tsetrand$ or $\tsetsplit$ appear to deviate significantly from their initial distribution, which corroborates well with their poor performance.
This distribution indicates that the best models are robust, in the sense of having similar parameters. Furthermore, training using information about fixed points appears equivalent to that based on a long trajectory. This observation is also quantified in terms of a radial distribution, see Fig.~\ref{Fig_radTSNE_0mem}. Both the models trained from $\tsetergo$ and $\tsetFP$ present a distribution with highest density around zero, while $\tsetrand$ and $\tsetsplit$ lead to a bias in the parameter distribution, hence being associated with poor models. 
In view of the entropy levels associated with each of the datasets (see Sec.~\ref{Sec_EntDS}), we conclude that organizing the datasets around the fixed points, and leveraging the simpler, linear dynamics characterizing the vicinity of those points corresponds to deploying a curriculum learning for the training of fully data-driven models. We postulate that a possible rationale behind the superior performance can indeed be understood by a simpler landscape of the cost associated with the data obtained where the dynamics is well approximated by linear behaviour. On the other hand, the curriculum learning assumption does not fully explain the discrepancies arising between models that -- in principle -- are qualitatively the same (namely, $\tsetergo$ and $\tsetsplit$). For this reason, we turn our attention to a crucial element: the role of the memory in the recurrent models.

\subsection{Handling the memory}\label{Sec_memory} Memory is considered in this section. Instead of the null memory (initialized to zero) considered so far and classically found in many applications, an initial memory chosen at random, drawn from a multivariate Gaussian distribution $\mathcal{N}(\boldsymbol{0}, \mathbf{I})$ is used. Compared to the null initial memory, the performances are seen to significantly increase for all models, as evidenced in Fig.~\ref{fig_rand_mem}. This is consistent with the fact that a non-zero hidden state improves the numerical conditioning of the training steps and prevents degenerate gradients, \cite{Ghazi_etal_19}.
The contrast in performance between models learned from $\tsetergo$ and $\tsetsplit$, which rely on the \emph{same} information, leads to identify a key point. The LSTM model trained with $\tsetergo$ learns to best fit one long trajectory. In the case of $\tsetsplit$, it learns to best fit 9 short trajectories, which however, when stitched together, match the $\tsetergo$ long trajectory. Yet, their performances are consistently very different in the null initialization situation, emphasizing that the initial memory of the retained recurrent network (LSTM) strongly affects the resulting performance. In the case of $\tsetergo$, the impact of an incorrect initial memory is diluted within a long trajectory and weakly affects the learning quality. When short trajectories are considered instead, the relative impact of an incorrect initialization is stronger and can introduce a significant bias, hence a poor resulting model.
This observation is further supported by the model learned from $\tsetFP$. While it also relies on samples originating from $9$ distinct, non sequential, segments, its performance is good, as evidenced by Fig.~\ref{Fig_d2_0mem}. What is unique about this particular training set is that the samples are from segments of trajectories originating from (unstable) fixed points, thus associated with a locally linear dynamics. This feature allows the incorrect initial memory to be progressively wiped out in time without significantly affecting the predicted dynamics.

\begin{figure}
\centering
\includegraphics[width=.5\textwidth]{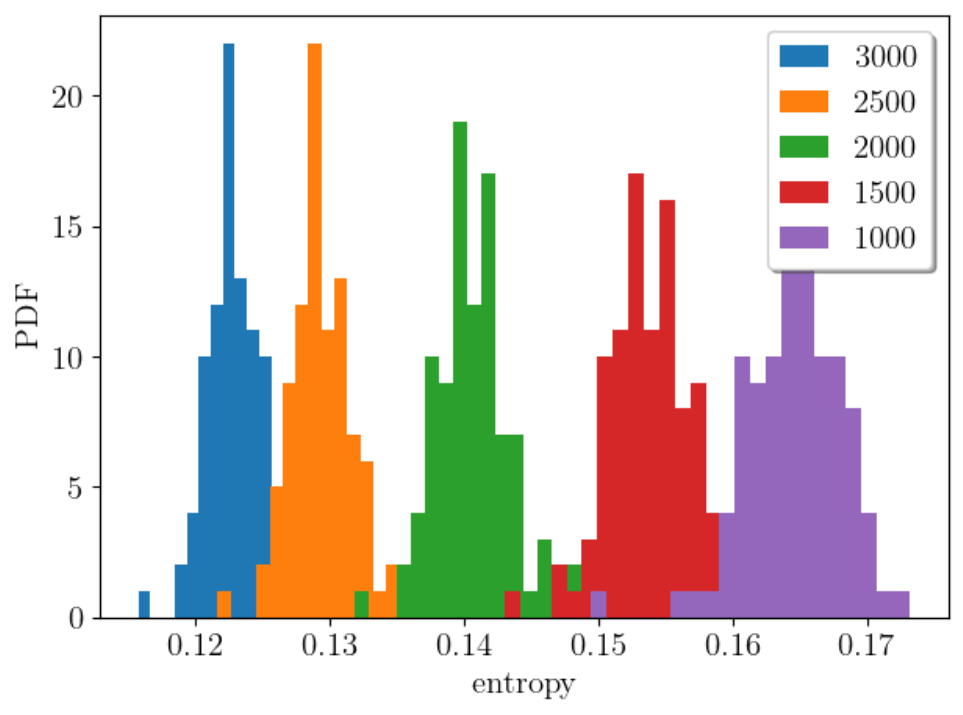}
\caption{Distribution of the SVD-entropy (Eq.~\ref{eq-svd}) associated with the structured dataset $\tsetFP$ (blue curve) as compared with the entropy computed for the same dataset shortened by progressively removing samples in the vicinity of the fixed points. The shortest trajectories are characterized by only $1000$ samples. The entropy progressively increases as the redundancy of the dataset reduces. The highest levels of entropy found for the latter case are comparable to the estimate computed for the reference dataset $\tsetergo$, as well as the cases $\tsetrand$ and $\tsetsplit$.}
\label{fig_shorting}
\end{figure}

\subsection{Can we use less data?}\label{Sec_shorter} 
\begin{figure*}
\centering
\includegraphics[width=.9\textwidth]{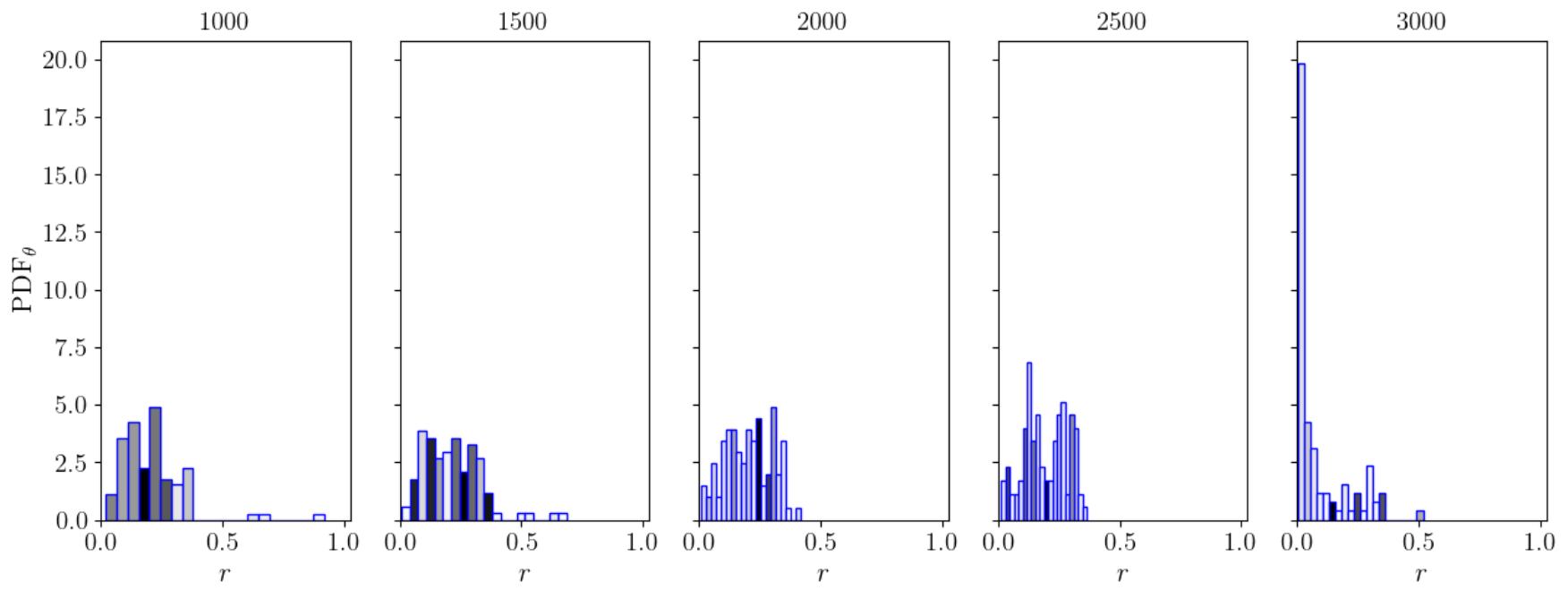}
\caption{Using the same postprocessing adopted in Fig.~\ref{fig_Zero_mem}, the radial distribution of the t-SNE projected model parameters is shown for the model based on the datasets obtained starting from $\tsetFP$, and shortened of the initial states. We order from light cyan to darker blue models performing progressively worse in terms of resulting $d_2$ accuracy; note that the level set of the blue gradient is normalized for each subplot. As a reference, we consider the case with $3000$ ($e$). A rather compact clustering of the models, \emph{i.e.} $r<0.5$, is found up to $t=2000$ (insets $c$ and $d$), suggesting good behavior of the trained models. Shorter trajectories lead to less robustness in terms of model weights as well as larger errors when considering $d_2$.}
\label{fig_shorter_rad}
\begin{picture}(0,0)
\put(-131,230){$(a)$}
\put(-51,230){$(b)$}
\put( 31,230){$(c)$}
\put(111,230){$(d)$}
\put(191,230){$(e)$} 
\end{picture}
\end{figure*}
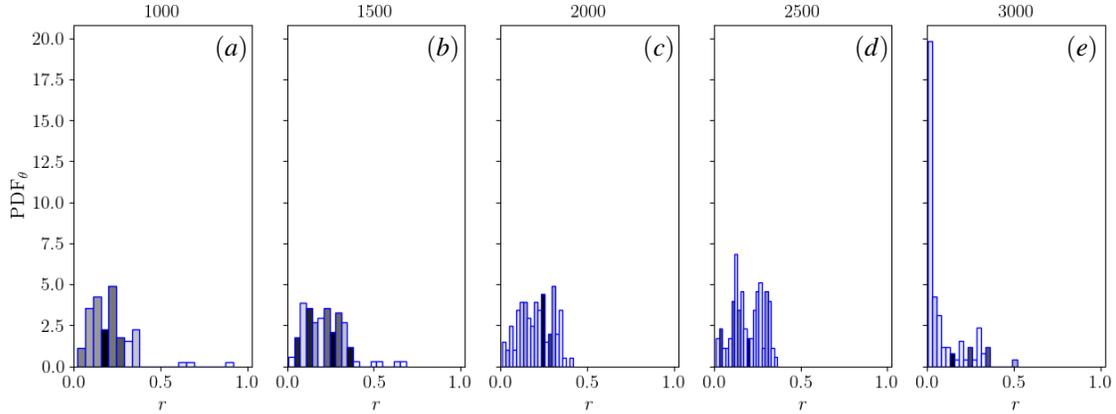

The results shown so far lead to slightly counter-intuitive conclusions: from one hand, low entropy indicates that the dynamics described by the datasets $\tsetFP$ lead to a simpler training; on the other hand, simpler dynamics means high predictability and thus less information. In that sense, one might wonder whether is possible to leverage these considerations for reducing the amount of data used during the training: in fact, the linear regions are less informative as they are redundant, in particular when we consider ``slow'' trajectories emanating from the fixed points located in the two lobes of the Lorenz'63 attractor. In order to provide more effective datasets, while preserving the curriculum learning idea, we consider here shortened trajectories where the initial samples are removed. The resulting characterization of the shortened datasets, based on the SVD entropy, is contained in Fig.~\ref{fig_shorting}: the entropy increases progressively as the dataset contains (relatively) more data stack in the nonlinear region. By considering these shortened datasets, we train LSTM models and assess their robustness using the $d_2$ metric applied in the previous sections, and study the associated models. As shown in Fig.~\ref{fig_shorter_rad}, it is found that we can reduce the number of samples to a total of $9 \times 2,000 = 18,000$, as the behavior is essentially the same as for the original $\tsetFP$ models, \emph{i.e.} not prone to overfitting and with low probability to produce biased results. When further reducing the data, results start to worsen until they definitively deteriorate for the last case $9 \times 1,000 = 9,000$. Nonetheless, we can conclude that a reduction of $33\%$ of the samples does not lead to loss of performance for the analyzed case. While the focus of the analysis is on data selection and their impact on the quality of performance, it is worth noting that any constraint applied to the cost function can further reduce the amount of data while preserving the performance of the system, for instance by explicitly introducing information about the stability of the system (Lyapunov exponents or the Jacobian matrix) or introducing multi-objective cost functions \cite{quade2016prediction, la2019probabilistic}. In that sense, a combination of physics constraints, data selection/distillation and architecture optimization can deeply improve the prediction abilities of data-driven models with a significant reduction of the data necessary for training.

\section{Concluding discussion}\label{Sec_discussion} In this work, we apply the concept of curriculum learning for data-driven modeling of a complex dynamical system: this concept is inspired from human behavior and is based on the idea that better learning is achieved when machines are trained using data of increasing complexity. Among the possible explanations for motivating the effectiveness of this learning strategy, a possible one is related to the shape of the cost landscape, that evolves from being near-convex to more complex as the complexity of the data increase. By sorting the data in this way, the optimization process is favored towards better extrema. Motivated by this methodology, we have performed a thorough analysis of some aspects inherent to the process of learning a complex physical system from observational data using the classical Lorenz'63, owing to its simplicity and hence the possibility to carry out an exhaustive campaign of simulations. As a learning method, we have considered LSTM neural networks, an established recurrent structure often used in such a context. We addressed the following issues:
\benum
\item Is there a minimum amount of data needed to obtain a robust model able to generalize?
\item Is it possible to go beyond this limitation thanks to some ``data-based'' knowledge of the system using a metric?
\item What is the impact of the initial state of the retained model on its resulting performance?
\eenum
The first two questions are rather fundamental. In particular, the first question concerns the amount of data required for training and it is related to what is known for classical embedding approaches, which is bound to fail when time-series data are not enough to reconstruct the entire phase-space. Generally speaking, in absence of prior information on the system at hand besides an estimated upper bound of its dimension, the ergodic theory, and notably the Kac lemma, provides a reliable criterion on the minimal amount of data necessary to train a model. This amount of data grows exponentially with the dimension of the attractor. Models trained on a smaller amount of training data have been shown to consistently lead to poor performance, as illustrated both via the time-series prediction and the resulting model estimated dimension assessment criteria. Our analysis hence suggests that recent results on dynamical systems showing a larger predictability horizon appear more as a result of overfitting than an intrinsic achievement of neural networks.
The second question is addressed by comparing different training sets. In particular, a training set acquired from a system initially close to one of its unstable fixed points exhibits a low entropy, low complexity, behavior owing to the essentially linear local regime, before reaching a higher complexity, fully nonlinear, regime when getting closer to the attractor. This essentially allows to disentangle the contribution of the different regimes of the underlying system and to sequentially focus the learning effort on these different regimes from specific parts of the training set, effectively resulting in a curriculum. Interestingly, shortening the trajectories emanating from the fixed points and following insights based on the entropy contents lead to results consistent with the ones obtained with a larger amount of data.
The last issue is related to the objective of introducing best-practice guidelines for practitioners. The numerical test provides evidences on the effectiveness of curriculum learning, as this choice leads to consistent and accurate models even with a limited amount of data with respect to what is prescribed by ergodic theory. Moreover, the analyzed experiments have neatly shown that starting from the fixed points of the dynamical system constitutes a workaround in learning the dynamics of the system from situations where the memory is essentially harmless to the learning process. Indeed, in these memory-based models, flushing an initially incorrect memory is crucial for obtaining a relevant and faithful model able to generalize its prediction beyond the sole situations encountered during the training step. Our numerical experiments clearly show that there is an impact of the initialization of the memory on the quality of the predictions provided by the LSTM models and it is particularly severe if one learns from a set of short-termed observations. Superior performance was consistently achieved with a memory initialized at random to avoid bias.  
While we here focused on the Lorenz'63 system and LSTM models, we believe these findings to be widely applicable and to provide some evidence-based good practice for data-driven modeling of physical systems. Future work will focus on a finer analysis of the interplay between the structure of the dynamical system and the learning process, as well as on the choice of appropriate metrics complementing entropy levels, toward a principled learning strategy. In that sense, curriculum learning strategies can be envisioned as a valuable tool for the analysis of complex dynamical systems and open up avenues for research within the context of active learning. 

\section*{Acknowledgement} 
This work was funded by the French Agence Nationale de la Recherche via the \texttt{Flowcon} project (ANR-17-ASTR-0022) and the \texttt{Speed} project (ANR-20-CE23-0025-01). L.M. gratefully acknowledges stimulating discussions with Alex Gorodetsky (University of Michigan, US). O.S. thanks Luca De Cicco (Politecnico di Bari, Italy) for exchanges on the role of entropy metrics in curriculum learning. S.C. gratefully acknowledge fruitful discussions with Angelo Vulpiani (University La Sapienza, Italy). 


\bibliographystyle{unsrtnat}
\bibliography{biblio}

\end{document}